\documentclass[letterpaper, 10 pt, conference]{ieeeconf}  

\IEEEoverridecommandlockouts                              

\usepackage{graphics} 
\usepackage{subfigure} 
\usepackage{epsfig} 
\usepackage{mathptmx} 
\usepackage{times} 
\usepackage{amsmath} 
\usepackage{amssymb}  
\usepackage{booktabs}
\usepackage{cite}

\title{\LARGE \bf
Collaborative Recognition of Feasible Region with Aerial and Ground Robots through DPCN
}

\author{Yunshuang Li$^{1}$, Zheyuan Huang$^{1}$, Zexi Chen$^{1}$, Yue Wang$^{1}$, Rong xiong$^{1,\dagger}$
\thanks{$^{1}$Yunshuang Li, Zheyuan Huang, Zexi Chen, Yue Wang and Rong Xiong are with the State Key Laboratory of Industrial Control Technology and
Institute of Cyber-Systems and Control, Zhejiang University, Zhejiang, China}%
\thanks{$^{\dagger}$Corresponding author, Email Address:
        {\tt\small  rxiong@zju.edu.cn}}%
\thanks{Funded by the National Key R${\&}$D Program of China(2018YFB1309300)}%
}

\begin{document}

\maketitle
\thispagestyle{empty}
\pagestyle{empty}

\begin{abstract}

Ground robots always get collision in that only if they get close to the obstacles, can they sense the danger and take actions, which is usually too late to avoid the crash, causing severe damage to the robots. To address this issue, we present a collaborative system with aerial and ground robots to gain precise recognition of feasible region. Taking the aerial robots' advantages of having large scale variance of view points of the same route which the ground robots is on, the collaboration work provides global information of road segmentation for the ground robot, thus enabling it to obtain feasible region and adjust its pose ahead of time. Under normal circumstance, the transformation between these two devices can be obtained by GPS yet with much error, directly causing inferior influence on recognition of feasible region. Thereby, we utilize the state-of-the-art research achievements in matching heterogeneous sensor measurements called deep phase correlation network(DPCN), which has excellent performance on heterogeneous mapping, to refine the transformation. The network is light-weighted and promising for better generalization. We use Aero-Ground dataset which consists of heterogeneous sensor images and aerial road segmentation images. The results show that our collaborative system has great accuracy, speed and stability.

\end{abstract}

\section{INTRODUCTION}

As the rapid development of robot operation system with vast deep research into it, the great possibility to  combine complementary information from different devices is dug recently, in order to increase the stability and safety of the system. The commonest cooperation is done by ground robots, such as mobile car, and aerial robots, like micro-aerial vehicles. 

As mentioned in \cite{7784317}, ground robots usually carry substantial payloads and actively interact with the environment. However, the operator can only receive limited information about its surroundings because of its low viewpoint. Drones, on the contrary, help to provide a situational assessment of the environment with the ability to cover large areas quickly. The data of global road segmentation ensures better performance in driving tasks, especially with potentially unknown and challenging obstacles on the ground. For example, in automatic driving systems, working in a heterogeneous team of flying and ground robots enhances the capabilities of robots to support high level safe service. 

\begin{figure}[thpb]
        \centering
        \includegraphics[scale=0.234]{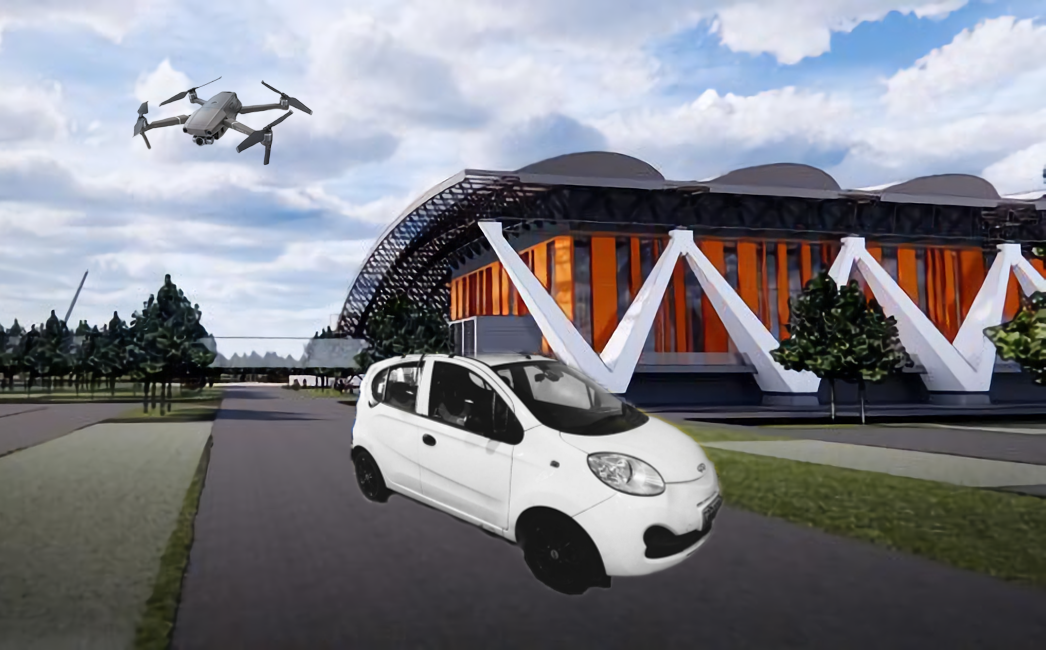}
        \caption{Demonstration of collaborative system carried by a
        ground robot, such as a car, and a drone with a camera on
        both of them.}
        \label{figurelabel}
     \end{figure}

Our collaborative system consists of a ground robot and a drone. The drone gives help to the ground robot by provding the images with road segmentation. As a consequence, the ground robot will get access to not only the global situation on the road but also the auxiliary recognition of fesible region, so that the ground robot performs better in interacting with the environment. Although the whole system consists of different devices to cooperate, our system still doesn't cost much time and keep real-time.


\subsection{Related work}

Prior work on collaborative system between ground robots and aerial robots pays close attention to completeing a guide system. The aerial robots guide the ground robots to a certain destination. Some scholars utilize visual markers \cite{enhance} or edge detection \cite{Robosys} to enable the aerial robots to track the ground robots. In addition, orthomosaic Map \cite{xu2020collaborative} can also apply to the localization problems for the ground robot in the global map given by the drone.

In our system, we focus on the information of road segmentation to help to obtain recognition of feasible region and avoid unnecessary collision. Road segmentation task can be done just with traditional image processing method. Many other methods are raised to improve the robustness of road segmentation, for example, a segmentation framework based on SVM learning \cite{inproceedings}. To deal with the disturb in the image, such as shadows, \cite{6}\cite{7} present effective approaches. Therefore, we skip the exhaustive process of segmentation temporaryly, supposing there has been preprocessing images from aerial robots. With these images, we only need a precise mapping using heterogeneous measurements.

For homogeneous image matching, there are a wide variety of mature methods. Part of them depend on  point features correspondences to localize in specific setups \cite{8}
while others utilize dense correlation methods to find the best pose candidate in solution space \cite{article9}. However, all these approaches cannot satisfy the demand 
of mapping with  heterogeneous images. When it comes to heterogeneous image matching, some scholars apply hand-craft features \cite{article10}  to realize localization tasks, yet the property of the frameworks has large influence on the coverage of the hand-crafted features, especially over complex terrain in real environment. According to the character of  heterogeneous image matching, it's convincing to use learning-based methods with certain generalization ability. \cite{8954371}\cite{8957240} learn the embeddings for heterogeneous observations and exhaustively search for the optimal pose in the discrete solution space. However, the universal problem of learning-based methods is its exhaustive evaluation on the large pose space. It suffers from low efficiency and cannot meet the demand of speed in the collaborative system.

What we need is to obtain an ideal learning-based mapping method, which can get the solution without exhaustive evaluation and also have good interpretability and generalization. 
In \cite{chen2020deep} and partial content in \cite{9361304}, scholars propose such a learnable matcher, of which the essence is a differentiable phase correlation. Specifically, they adopt the conventional phase correlation pipeline and explicitly endow the Discrete Fourier Transform (DFT) layer, log-polar transformation layer (LPT). 
That's because phase correlation is a similarity-based matcher that performs well for inputs with the same modality yet only tolerate small high frequency noise. The first two layers have the ability to deal with this problem. Then they apply  differentiable correlation layer (DC). 
By modifying traditional phase correlation  into a differentiable manner, the DC layer becomes trainable and is embedded into the  end-to-end matching network. This architecture is shown in Figure 3. The whole network is called deep phase correlation network(DPCN). It has a vast variety of applictions, such as PRoGAN \cite{chen2020pose}. In this paper, scholars utilize DPCN as a crucial part of their network, which is utilized for weakly paired image style translation. 

In our system, we take the advantages of DPCN to improve the mapping results. Aerial robots get corresponding images according to the information of GPS, then complete the preprocessing of road segmentation with global viewpoints, which contain the initial value of transformation to the ground robots' images.
With the DPCN, we recalculate the value of transformation by converting these images to the viewpoints of ground robots, which is heteroid from the original images. As a result, ground robots get the information of road segmentation in the view of aerial robots, thus obtaining the feasible region much more precisely and interacting with the environments neatly and quickly. 
In our experiments, we train DPCN to accomplish the recomputation of the transformation, and show the robustness and efficiency in the proposed system.

\subsection{Paper outline}

Our goal is to convert the preprocessing image with road segmentation from aerial robots to the viewpoints of  ground robots, so that the ground robots get the road segmentation information to interact better with the environment. Therefore, we present a collaborative system to complete this task.



We describe the construction and the implementation details of the whole system in Sec. II. The details of the dataset, along with the experiments process are described in Sec. III. The results in Sec. IV show that our collaborative system can perfectly satisfy the demand of accuracy and instantaneity. Finally, the work is concluded in Sec. V.

\section{COLLABORATIVE SYSTEM}

The work flow of the collaborative system is shown in Figure 2. It intuitively demonstrated the collaboration carried between the ground robot and a drone.

\begin{figure}[thpb]
        \centering
        \includegraphics[scale=0.33]{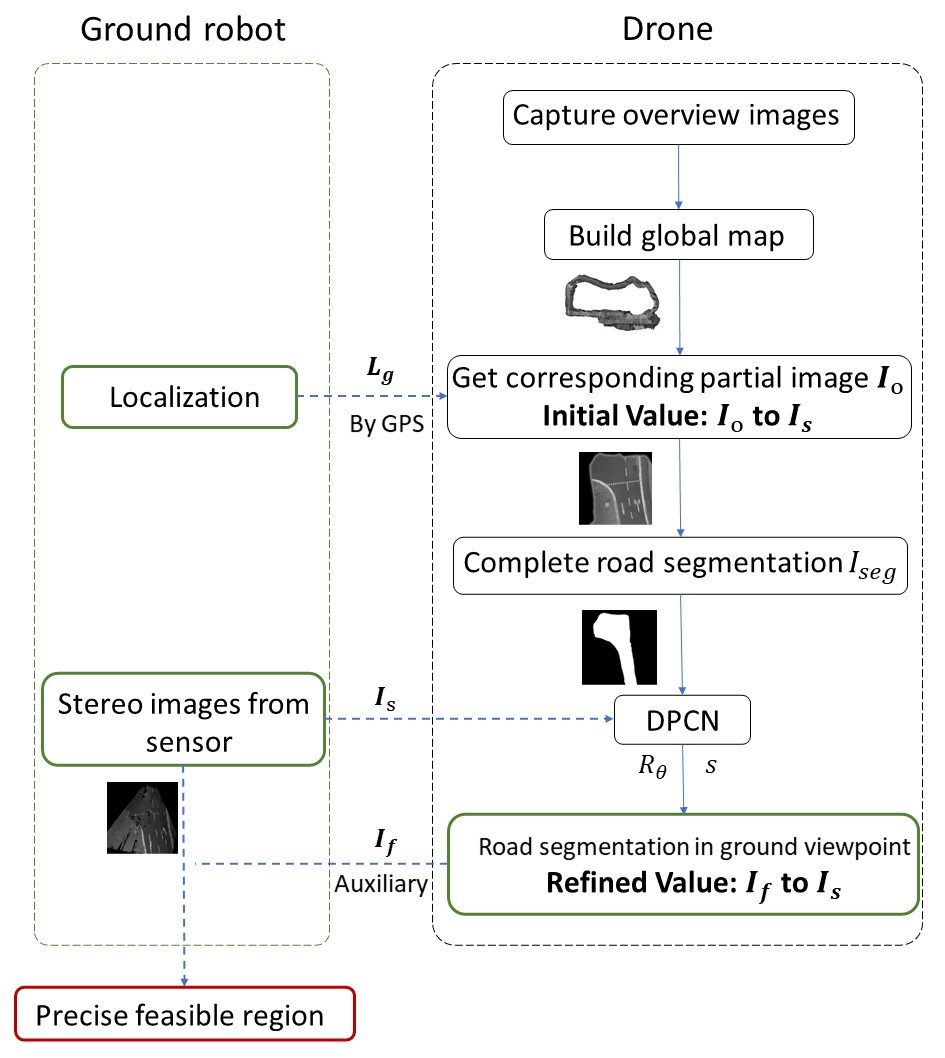}
        \caption{The work flow of the collaborative system. Green blocks mean that images are in the viewpoint of ground robots while black blocks are that of a drone. Blue dotted lines refer to the communication in the collaborative system.}
        \label{figurelabel}
     \end{figure}

The pseudocode shown in TABLE I and II explains how the collaborative work.

\begin{table}[h]
        \caption{PSEUDOCODE OF THE COLLABORATIVE SYSTEM}
        \begin{center}
        \begin{tabular}{l}
        \toprule
        \textbf{Algorithm 1} Framework of the drone for our collaborative system\\
        \midrule
        \textbf{Input:} The ground robot's localization $L_g$ done by GPS and sensor \\
        image from ground robot $I_s$.\\
        \\
        \textbf{Output:} The corresponding image $I_f$ located at $L_g$ with road segmen-\\
        tation information in the viewpoint of the ground robot.\\
        \\
        \quad1: ... // Capture overview images, then build global map\\
        \quad2: \textbf{for} each $i \varepsilon$ [sections of global map] \textbf{do}\quad//Get initial value\\
        \quad3: \quad\quad \textbf{if} location at i match that of $L_g$ \textbf{then}\\
        \quad4: \quad\quad\quad\quad $I_o$ = i\\
        \quad5: \quad\quad\textbf{end}\\
        \quad6: \textbf{end}\\
        \quad7: $I_{seg}$ = Road Segmentation($I_o$)\quad//Complete road segmentation\\
        \quad8: $R_\theta, s$= DPCN($I_{seg}$, $I_s$)\quad//Feed into DPCN and return $R_\theta $ and $s$\\
        \quad9: $I_f$ = Image Transformation($I_{seg}$, $S$)\quad//Get refined value\\
        \;\;10: Send Image($I_f$)\\
        \bottomrule
        \end{tabular}
        \end{center}
        \end{table}

\begin{table}[h]
        \caption{PSEUDOCODE OF THE COLLABORATIVE SYSTEM}
        \begin{center}
        \begin{tabular}{l}
        \toprule
        \textbf{Algorithm 2} Framework of the ground robot for our collaborative \\
        system\\
        \midrule
        \textbf{Input:} The corresponding image $I_f$ located at $L_g$ with road segmenta-\\
        tion information in the viewpoint of the ground robot.\\
        \\
        \textbf{Output:} The ground robot's localization $L_g$ and the feasible region for\\
        ground robot $R_{fea}$\\
        \\
        \quad1: Send Localization($L_g$)\\
        \quad2: Send Sensor Image($I_s$)\\
        \quad3: Receive Image($I_f$)\\
        \quad4: Obtain Feasible Region($I_s$, $I_f$)\\
        \bottomrule
        \end{tabular}
        \end{center}
        \end{table}
Under such concise structure, collaboration between the ground robot and the drone can be done efficiently and fleetly with brilliant effect on the ground robot's sensing. Among all the steps in collaborative system,
it's of great significance to recalculate the transformation from the drone's images to the ground robot's images. Thereby, we introduce DPCN to complete this pivotal task.

In DPCN, in order to describe the transformation from the birds-eye view of aerial robots to the ground robots, scholars define a similarity transform S:

\begin{equation}
S = \left(
        \begin{matrix}
        sR_\theta & t\\
        0 & 1
        \end{matrix}
    \right)\in \mathbb{S} \mathbb{I} \mathbb{M} \left(2\right)
\end{equation}
where $s\in\mathbb{R} ^+ $ is the scale, $R_\theta\in\mathbb{S} \mathbb{O} \left(2\right)$ is the rotation matrix generated by the heading angle $\theta$ ,and $t \in \mathbb{R} ^2$is the translation.
With the result of S, we can describe the mapping process as follows:

\begin{equation}
        I_f = T(f(I_o),S)
\end{equation}
$I_o$ refers to the original image, while $I_f$ refers to the final result with transformation $S$. $f$ and $T$ means certain operation with the image, aiming to achieve better mapping results.

The deep phase correlation network consists of three transformation, the fast Fourier transform(FFT), the log-polar transform(LPT) and the differentiable phase correlation solver(DC). Images are transferred into Fourier frequency domain with invariance of translation in FFT. Thereby, rotation and scale are represented by frequency and magnitude. In log-polar transform, we transfer the data included in the frequency domain from cartesian to log-polar coordinate system,
making it solvable with next operation. 
Through using expectation as estimation of DC, phase correlation becomes differentiable, therefore learnable, and eventually output the results of the transformation between two inputs. When learning, a temperature coefficient is assigned to tune the range of feature input, which accelerates the convergence.

\begin{figure}[thpb]
        \centering
        \includegraphics[scale=0.34]{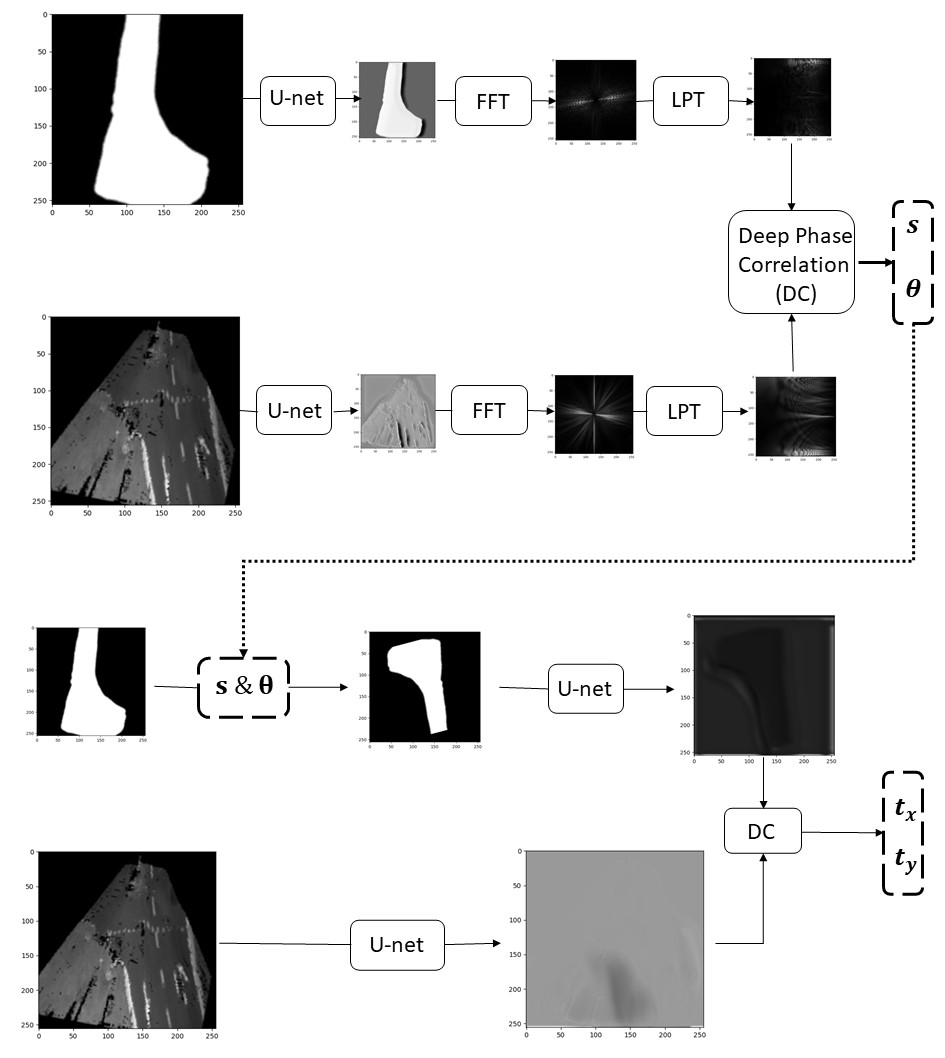}
        \caption{The structure of the pipeline with U-net, fourier transformation layer(FFT), log-polar transformation layer(LPT) and deep phase correlation layer(DC). We can get the result of rotation angle and scale, then apply them to calculate transformation in x and y.}
        \label{figurelabel}
     \end{figure}

\begin{figure}[thpb]
        \centering
        \includegraphics[scale=0.34]{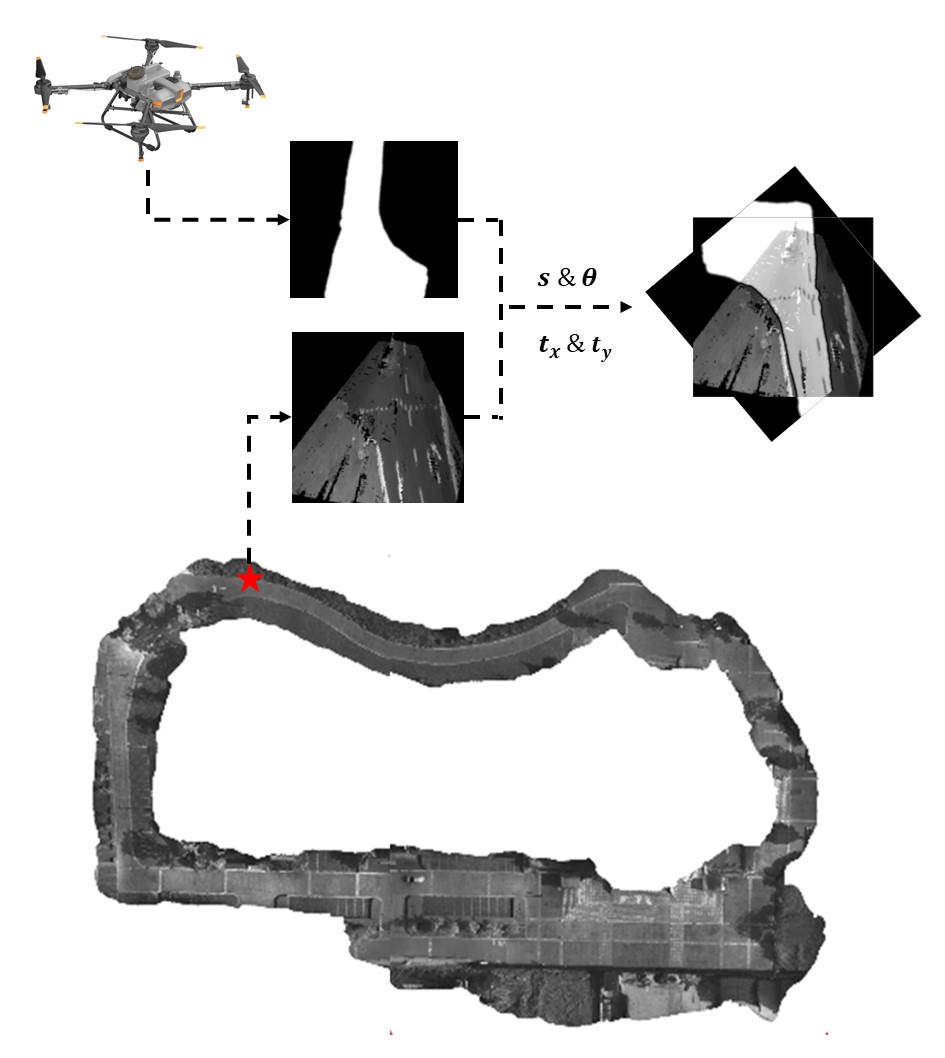}
        \caption{We utilize the result of the pipeline to complete our collaborative system. The picture shows the overview of our collaborative system.}
        \label{figurelabel}
     \end{figure}

We utilize DPCN to obtain $S$ from drone's viewpoint to that of the ground robot and carry out the transformation. To get the result of $s$ and $R_\theta$, we design such pipeline shown in Figure 3. Paired images are input to the deep phase correlation network. After the operation of a feature extractor and three well designed layers, $s$ and $\theta$ have been calculated. Then we utilize the results to preprocess the input images in transformation section.
FFT and LPT have no difference in translation, therefore only a feature extractor is adequate to calculate $t_x$ and $t_y$.

This method can help eliminate the exhaustive evaluation and keep high speed in calculation, ensuring our collaborative system to work smoothly.
\section{EXPERIMENTS: DATASET AND SETUP}

We apply the method of DPCN to refine the result of transformation between heterogeneous images. In our experiments, we'd like to take the advantages of the images with road segmentation in the view of the aerial robots to assist the ground robots to perform better in interacting with the environment. Therefore, we set the stereo images from ground robots as template, while the drone's view images as source. After training the DPCN, we can get the refined value of transformation.
We compare the initial value of transformation by GPS and the refined value by DPCN to show the preponderance of our work. This operation is of widely benefits to help ground robots to avoid collilsions into obstacles.

\subsection{Dataset} 

To achieve the best results of our collaborative system, the Aerial-ground dataset \cite{AGDataset} can be a brilliant option. The open dataset has four types of heterogeneous images, containing “drone's view”, “LiDAR intensity”, “stereo” and “satellite”. Firstly, we get corresponding image (b) in Figure.3 from global map (a) by GPS, then conduct road segmentation tasks on the drone's view shown in (c). The transformation from (c) to stereo images from ground robots' sensor (d) is defined as initial value. In DPCN, stereo images are set as template, while drone's view images with road segmentation are set as source. Through deep phase correlation network, the mapping between this two types of images is refined. The groundtruth of this pair focuses on sacle and rotation,
thus we only evaluate the network of these two parameters.

\begin{figure}[thpb]
        \centering
        \subfigure[The overview of the whole open dataset]{
        \label{fig:subfig:a} 
        \includegraphics[width=3.2in]{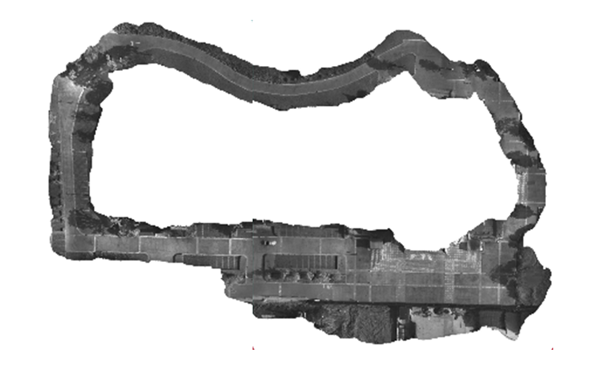}}
        \\
        \subfigure[Drone's view]{
        \label{fig:subfig:b} 
        \includegraphics[width=0.99in]{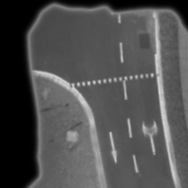}}
        \subfigure[Drone's view with road segmentation]{
        \label{fig:subfig:c} 
        \includegraphics[width=0.99in]{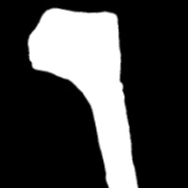}}
        \subfigure[Stereo images of ground robots’ view]{
        \label{fig:subfig:d} 
        \includegraphics[width=0.99in]{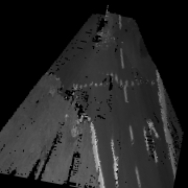}}
        \caption{All images in the open data set are sections of (a), which is taken by the collaborative system. (b) is original drone's view image. (c) and (d) demonstrate each part of the dataset used for training--one is from drone's view with road segmentation work, the other one is stereo image with the viewpoints of MAV on the ground.}
        \label{figurelabel}
\end{figure}

\subsection{Setup}
\textbf{Hardware:} The experiments are all conducted on a single laptop (Intel® Core™ i7-9750H CPU @ 2.60GHz $\times$ 12) with the GPU of an RTX 2060.

\textbf{Software:} The network is trained under 30 epoches each with two batch size. 3640 pairs of imges is prepared by the experiments. 2/3 of the images are used for the training part, while the others are used to validate the network.

\section{EXPERIMENTS: RESULT}

Based on the current setup of hardware and software, we carried out the experiments. It cost almost 12 hours to complete the whole training process, with the loss convergencing and the model becoming suitble for our collaborative systems.

In order not to make the model overfitting, we kept an eye on the process of training elaboratively on the Tensor board. As the loss getting smaller and the accuracy of scale and rotation rising, we deem the model performs well for the collaborative system and stop training.

There are several parameters of the network after training shown in TABLE III. By the relatively low learning rate and best loss, it's explicitly revealed that our model has been trained suitable to complete the mapping from source to template.
\begin{table}[h]
\caption{RESULTS OF TRAINING}
\begin{center}
\begin{tabular}{|c|c|c|}
\hline
Epoch & Lr & Beat loss \\
\hline
30/30 & 6.18e-06 & 12.6787\\
\hline
\end{tabular}
\end{center}
\end{table}

Then, it's time to conduct the validation experiments on the network to varify its accuracy and speed. We set two experiments with different amount of images as validation, thus ensuring the reliability and universality of the application into DPCN.

The model is evaluated with the accuracy of rotation and scale as well as the time it consumes for one time on average. The calculation of the accuracy is defined as follows:
\begin{equation}
        Acc_{scale} = \frac{1}{n}\sum_{i = 1}^{n}(1-|\frac{s-s^*}{s^*}|) 
\end{equation}

\begin{equation}      
        Acc_{rot} = \frac{1}{n}\sum_{i = 1}^{n}(1-|\frac{\theta -\theta ^*}{\theta ^*}|)   
\end{equation}
where $s$ and $\theta$ refers to the results of the calculation in DPCN, $s^*$ and $\theta^*$ refer to the groundtruth of scale and rotation.
$n$ denotes the amount of images in validation. 

\begin{table}[h]
\caption{RESULTS OF VALIDATING}
\begin{center}
\begin{tabular}{|c|c|c|c|c|c|}
\hline
Scale of validation set& $Acc_{scale}$ & $Acc_{rot}$ & $Acc_{x/y}$ & Time \\
\hline
3640 pairs & 83.20$\%$ & 94.07$\%$ &100.00$\%$ & 77.61ms\\
\hline
910 pairs& 80.11$\%$ & 89.51$\%$ & 100.00$\%$ &76.59ms \\
\hline
\end{tabular}
\end{center}
\end{table}

\begin{table}[h]
        \caption{COMPARISION BETWEEN INITIAL VALUE AND REFINED VALUE FOCUSED ON ROTATION}
        \begin{center}
        \begin{tabular}{|c|c|c|}
        \hline
        Initial value & Refined value & Improvement \\
        \hline
        89.41 degree & 5.30 degree & 94.07$\%$  \\
        \hline
        \end{tabular}
        \end{center}
        \end{table}

Here comes the results of validation. It is demonstrated in TABLE IV that we get relatively good network model to achieve our goal of auxiliary road segmentation task on ground robots. Due to the speciality of the open dataset, we concentrate on the accuracy of scale and rotation. It's deserved to be mentioned that both of the accuracy is over 80$\%$,
with the rotation accuracy reaching even about 90$\%$. We are rather delighted to notice that our model consumes only about 77.61ms and 76.59ms for one calculation. We compare the transformation before and after DPCN and show the improvement in TABLE V, which contributes a lot to the concise feasible region recognition.

We choose two of the pair images to show the final results of our collaborative system:
\begin{figure}[thpb]
        \centering
        \includegraphics[scale=0.24]{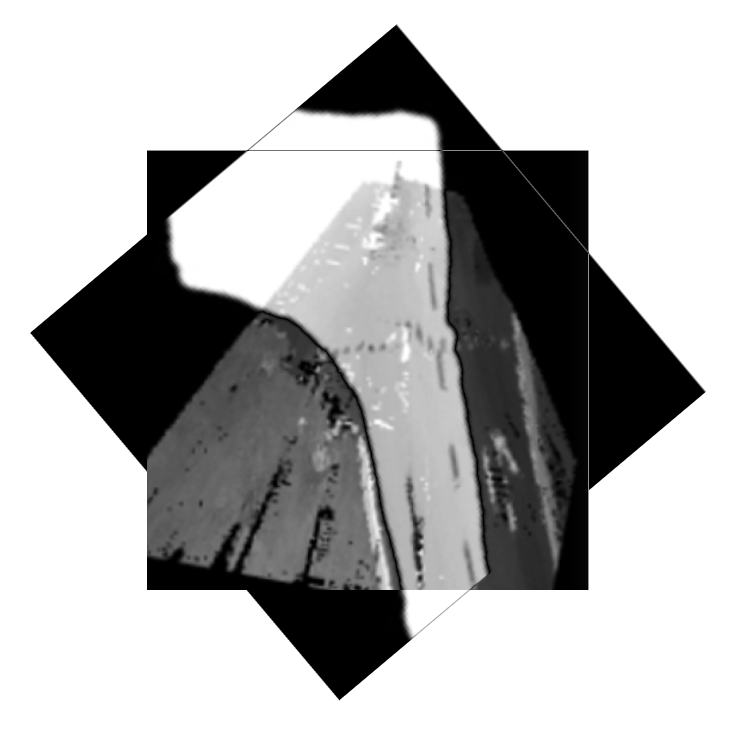}
        \caption{$\theta$ = 41.4844 degree, $s$ = 1.0394 with its groundtruth $\theta^*$ = 41.6478 degree, $s^*$ = 1.1537 }
        \label{figurelabel}
     \end{figure}

\begin{figure}[thpb]
        \centering
        \includegraphics[scale=0.24]{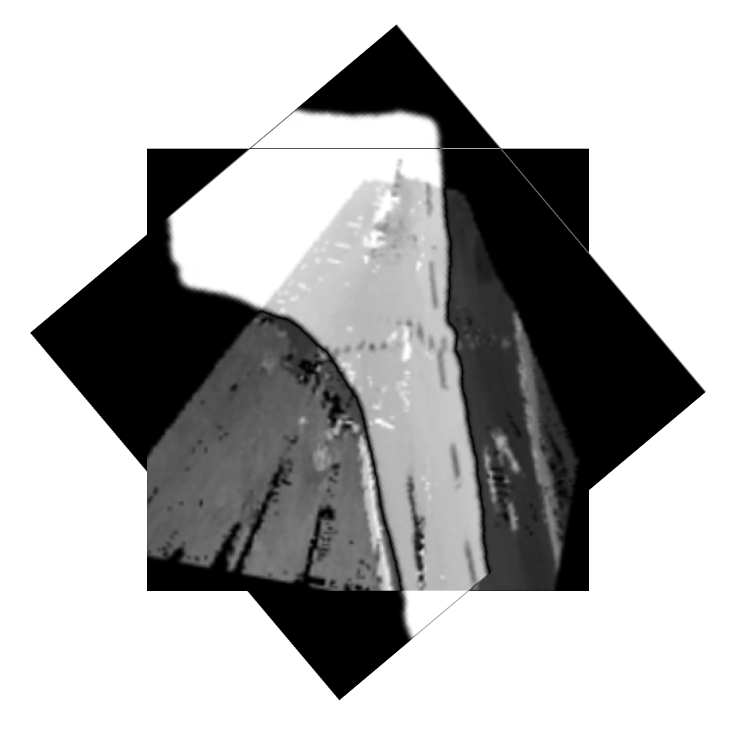}
        \caption{$\theta$ = 49.9219 degree, $s$ = 1.0394 with its groundtruth $\theta^*$ = 51.8730 degree, $s^*$ = 1.1860 }
        \label{figurelabel}
     \end{figure}

\section{CONCLUSIONS}
We present a collaborative system with the application of deep phase correlation network. Aiming to assist ground robots to obtain feasible region and perform better in interacting with the environment, it's of great use to take advantages of a collaborative system with aerial robots' help. We utilize DPCN to improve the accuracy of mapping of road segmentation imges from aerial viewpoints to the ground. Our model 
has high accuracy of mapping so that the road segmentation information can be translated precisely to the ground viewpoints to help ground robots avoid obstacles. In the meantime, it's capable of running in real-time to raise the efficiency of the collaborative system.

\addtolength{\textheight}{-12cm}   









\bibliographystyle{unsrt}
\bibliography{reference}

\end{document}